\documentclass[%
 aip,
 amsmath,amssymb,
 reprint,%
]{revtex4-1}

\usepackage{subcaption}

\usepackage{tikz}
\usepackage{tikz-cd}
\usepackage{pgfplots}
\usepackage{hyperref}
\usepackage{adjustbox}
\usepackage{makecell}
\usepackage[capitalize,nameinlink]{cleveref}
\Crefname{section}{Sec.}{Secs.}

\usepackage{todonotes}
\presetkeys%
    {todonotes}%
    {inline}{}

\makeatletter
\def\@email#1#2{%
 \endgroup
 \patchcmd{\titleblock@produce}
  {\frontmatter@RRAPformat}
  {\frontmatter@RRAPformat{\produce@RRAP{*#1\href{mailto:#2}{#2}}}\frontmatter@RRAPformat}
  {}{}
}%
\makeatother

\usepackage{csquotes}

\usetikzlibrary{
  arrows.meta,                        %
  backgrounds,                        %
  ext.paths.ortho,                    %
  ext.positioning-plus,               %
  ext.node-families.shapes.geometric, %
  calc}                               %
\pgfplotsset{compat=newest}

\usepgfplotslibrary{patchplots}
\usepgfplotslibrary{fillbetween}
\pgfplotsset{%
    layers/standard/.define layer set={%
        background,axis background,axis grid,axis ticks,axis lines,axis tick labels,pre main,main,axis descriptions,axis foreground%
    }{
        grid style={/pgfplots/on layer=axis grid},%
        tick style={/pgfplots/on layer=axis ticks},%
        axis line style={/pgfplots/on layer=axis lines},%
        label style={/pgfplots/on layer=axis descriptions},%
        legend style={/pgfplots/on layer=axis descriptions},%
        title style={/pgfplots/on layer=axis descriptions},%
        colorbar style={/pgfplots/on layer=axis descriptions},%
        ticklabel style={/pgfplots/on layer=axis tick labels},%
        axis background@ style={/pgfplots/on layer=axis background},%
        3d box foreground style={/pgfplots/on layer=axis foreground},%
    },
}

\usepgfplotslibrary{groupplots}
\usepgfplotslibrary{polar}
\usepgfplotslibrary{smithchart}
\usepgfplotslibrary{statistics}
\usepgfplotslibrary{dateplot}
\usepgfplotslibrary{ternary}

\tikzset{
  basic box/.style={
    shape=rectangle, rounded corners, align=center, draw=#1, fill=#1!2, very thick},
  header node/.style={
    node family/width=header nodes, rounded corners,
    text depth=+.3ex, fill=white, draw},
  header/.style={%
    inner ysep=+1.5em,
    append after command={
      \pgfextra{\let\TikZlastnode\tikzlastnode}
      node [header node] (header-\TikZlastnode) at (\TikZlastnode.north) {#1}
      node [span=(\TikZlastnode)(header-\TikZlastnode)]
           at (fit bounding box) (h-\TikZlastnode) {}
    }
  },
}

\usepackage{xcolor}
\definecolor{onyx}{HTML}{393E41}
\definecolor{redviolet}{HTML}{8B2635}
\definecolor{munsellblue}{HTML}{4B88A2}
\definecolor{maxpurple}{HTML}{7E2E84}
\hypersetup{
    hidelinks,
    citecolor={green!50!black},
    urlcolor={blue!80!black}
}

\draft %

\begin{document}

\preprint{AIP/123-QED}

\title{Improving the Noise Estimation of Latent Neural Stochastic Differential Equations}

\author{L. Heck}
\altaffiliation[Also at ]{Department of Computer Science, RWTH Aachen University, Germany, and Complexity Science, Potsdam Institute for Climate Impact Research, Potsdam, Germany.}
\affiliation{
    Institute for Computing and Information Sciences, Radboud University, Nijmegen, the Netherlands
}

\author{M. Gelbrecht}
\altaffiliation[Also at ]{Complexity Science, Potsdam Institute for Climate Impact Research, Potsdam, Germany.}
\affiliation{
    Earth System Modelling, School of Engineering and Design, TU Munich, Germany
}

\author{M. Schaub}
\affiliation{
    Computational Network Science, Dept of Computer Science, RWTH Aachen University, Germany
}

\author{N. Boers}
\altaffiliation[Also at ]{Complexity Science, Potsdam Institute for Climate Impact Research, Potsdam, Germany, and Department of Mathematics and Statistics, University of Exeter, Exeter, United Kingdom.}
\affiliation{
    Earth System Modelling, School of Engineering and Design, TU Munich, Germany
}

\date{\today}%

\begin{abstract}
    Latent neural stochastic differential equations (SDEs) have recently emerged as a promising approach for learning generative models from stochastic time series data.
    However, they systematically underestimate the noise level inherent in such data,
    limiting their ability to capture stochastic dynamics accurately.
    We investigate this underestimation in detail and propose a straightforward solution:
    by including an explicit additional noise regularization in the loss function, we are able to learn a model that accurately captures the diffusion component of the data.
    We demonstrate our results on a conceptual model system that highlights the improved latent neural SDE's capability to model stochastic bistable dynamics.
\end{abstract}

\maketitle

\begin{quotation}
    The goal of fitting models to stochastic time series data is to match the data's probability
    distribution.
    While latent neural SDEs are a promising tool for this task,
    it has been observed that these models underestimate the noise level in the
    data.
    On complex stochastic systems such as multistable systems that are forced by
    noise, this underestimation can lead to a complete mismatch of behaviour
    between the model and the data.
    In this work, we analyse this underestimation and propose augmenting the
    training loss with an additional noise penalty.
    This penalty, adjustable via a hyperparameter, enables the model to align
    with user-defined target statistics. We demonstrate the effectiveness of our
    approach on a conceptual model system that exhibits bistable dynamics and is
    forced by noise. Our results show that the proposed noise penalty allows the
    latent neural SDE to accurately match the diffusion component of the data,
    improving the model's ability to fit the data distribution. The code for our
    experiments is publicly available.
\end{quotation}

\section{Introduction}%
Neural stochastic differential equations combine stochastic differential equations (SDEs) with a deep learning component. They have recently emerged as a powerful framework to model continuous data in an unsupervised, generative fashion.
Compared to standard (stochastic) recurrent neural networks, the underlying process is not a discrete-time process but a continuous SDE\@.
The model can thus be arbitrarily discretized and is amenable to direct analysis, thereby offering increased interpretability\@. 
Potential applications of neural SDEs cover a wide range of fields, including all instances in which SDEs are already used to investigate dynamical data, such as paleoclimatology and present-day climate \citep{hasselmann1976, franzke2015}, solar and wind energy forecasts \citep{moller2016, dong2020}, astrophysics \citep{fagin2023}, or finance \citep{black1973}. 

In general terms, such a neural SDE is defined by
\begin{equation}\label{eq:SDE}
  d u(t) = h(u(t), t) dt + g(u(t), t) d B_t,
\end{equation}
where $u(t)$ denotes the hidden state (vector) of the system, $B_t$ is standard Brownian motion and $h(\cdot)$ and $g(\cdot)$ denote the so-called \emph{drift} and the \emph{diffusion} term, respectively, which are parametrized via deep neural networks.
Two main methods have been proposed to fit the neural SDE to data: latent neural SDEs and SDE-Generative Adversarial Networks (GANs).
First, \emph{latent neural SDEs} \citep{li2020scalable} may be derived by minimizing the Kullback-Leibler divergence between the neural SDE and the empirically observed data $u_\text{obs}$; these can also be considered as a form of variational autoencoder~\cite{kidgerthesis}.
Second, \emph{SDE-GANs} \citep{kidgergan}, emerge from minimizing a Wasserstein type loss between $u$ and $u_\text{obs}$, which leads to a generative adversarial neural network architecture.
In this paper, we focus on latent neural SDE architectures, which are easier to train in practice, whereas SDE-GANs suffer both from the instability generally present in GANs and from a lack of known in-practice tricks to alleviate this instability~\citep[84]{kidgerthesis}.

Our main contributions are as follows. 
First, we provide a detailed investigation of the previously observed effect (\citealt{li2020scalable}, \citealt[p.\ 84]{kidgerthesis}) that latent neural SDE schemes are unable to accurately match the diffusion component of the data, resulting in failure to match the distribution of the data. 
To this end, we provide a detailed experimental evaluation of latent neural SDEs regarding this mismatch of the diffusion term and investigate the reasons for this effect.
Even though there have been experiments on latent neural SDEs in the literature (e.g., \citealt{li2020scalable}, \citealt[pp.\ 89--92]{kidgerthesis}, \citealt{kidgerefficient}, \citealt{zeng}), a detailed empirical investigation of this phenomenon is still missing.
These experiments aim at some form of \emph{uncertainty quantification}, where noise underestimation is more acceptable, while we aim at a model that can accurately reproduce the data distribution.

Using the newly gained understanding of latent neural SDEs, multiple solutions can be proposed to accurately match the diffusion component. As our second contribution, we provide a possible remedy in terms of a new noise penalty that we add to the loss function to be optimized.
We show that this enables the latent neural SDE to fit transition rates in multistable models and match the data's Kramers-Moyal (KM) coefficients, assuming constant diffusion in the data.
While we choose these metrics, any metric can be used to tune the noise penalty's hyperparameter.

Our experiments and all underlying data of the presented results are publicly available.\footnote{The artifact is available on Zenodo: \url{https://doi.org/10.5281/zenodo.14534738}.}

In the following, we first discuss related work in \cref{sec:relatedwork} and review latent neural SDEs in \cref{sec:latentSDEs}.
We then introduce the chosen experimental setups in \cref{sec:experiments}, before investigating the noise underestimation in \cref{sec:investigation}.
In \cref{sec:noise}, we present some approaches to fix the observed noise underestimation problem, before concluding with a short discussion of our results and possible future work.

\section{Related Work}\label{sec:relatedwork}
Neural SDEs can be used for uncertainty quantification within probabilistic machine learning, similarly to Gaussian Processes \citep{williams2006gaussian}, which also induce probability distributions on time series.
In contrast to Gaussian Processes, which are usually trained explicitly using a specified parametric kernel, we train neural SDEs using an unsupervised scheme.
The reason is that, in general, there is no way to write down the probability distribution that is induced by a neural SDE. 
The advantage of using SDEs instead of Gaussian Processes is that an SDE can express a much larger set of probability distributions than a Gaussian Process can.

Direct applications of latent neural SDEs include the following works from different disciplines.
\citealt{fagin2023latent} apply latent neural SDEs to data from astrophysics. They use them as a drop-in replacement for Gaussian Processes, so the noise is interpreted as uncertainty. \citealt{xu} consider latent neural SDEs with a fixed diffusion, a continuous analogue of Bayesian recurrent neural networks. \citealt{kidgerefficient} introduce a reversible Heun solver for differential equations and train a latent neural SDE on an air quality data set from Beijing. \citealt{hasan2021identifying} propose a different latent neural SDE framework to recover a mapping from one high-dimensional probabilistic space to another. Furthermore, the framework has also been extended to include discontinuous (`jump') processes by \citealt{jia2020neural}, which has been applied by \citealt{herrera2021neural} to predict mortality rate numbers in intensive care units. \citealt{physicsconstrained} learn an SDE diffusion term that is penalized on distance, and in contrast to our work, they use a custom training scheme tailored to their problem instead of one based on GANs or VAEs.

Aside from neural SDEs, there are many other approaches to data-driven modelling of stochastic time series. These include other generative models, such as denoising diffusion models \cite{finzi2023user} and GANs \cite{gagne2020machine}. Reservoir Computing, a deterministic, recurrent neural network, has been adapted as well to stochastic time series. However, these approaches usually have to include include additional steps, such as filtering \cite{jinno2025long} or past noise forecasting \cite{hasan2021identifying}. 

\section{Latent neural SDEs}\label{sec:latentSDEs}

\newsavebox{\dataplot}
\sbox{\dataplot}{%

\begin{tikzpicture}[/tikz/background rectangle/.style={fill={rgb,1:red,1.0;green,1.0;blue,1.0}, fill opacity={0.0}, draw opacity={0.0}}, show background rectangle]
\begin{axis}[point meta max={nan}, point meta min={nan}, legend cell align={left}, legend columns={1}, title={}, title style={at={{(0.5,1)}}, anchor={south}, font={{\fontsize{14 pt}{18.2 pt}\selectfont}}, color={rgb,1:red,0.0;green,0.0;blue,0.0}, draw opacity={1.0}, rotate={0.0}, align={center}}, legend style={color={rgb,1:red,0.0;green,0.0;blue,0.0}, draw opacity={1.0}, line width={1}, solid, fill={rgb,1:red,1.0;green,1.0;blue,1.0}, fill opacity={0.0}, text opacity={1.0}, font={{\fontsize{8 pt}{10.4 pt}\selectfont}}, text={rgb,1:red,0.0;green,0.0;blue,0.0}, cells={anchor={center}}, at={(1.02, 1)}, anchor={north west}}, axis background/.style={fill={rgb,1:red,1.0;green,1.0;blue,1.0}, opacity={0.0}}, anchor={north west}, xshift={1.0mm}, yshift={-1.0mm}, width={125.0mm}, height={61.5mm}, scaled x ticks={false}, xlabel={}, x tick style={color={rgb,1:red,0.0;green,0.0;blue,0.0}, opacity={1.0}}, x tick label style={color={rgb,1:red,0.0;green,0.0;blue,0.0}, opacity={1.0}, rotate={0}}, xlabel style={at={(ticklabel cs:0.5)}, anchor=near ticklabel, at={{(ticklabel cs:0.5)}}, anchor={near ticklabel}, font={{\fontsize{11 pt}{14.3 pt}\selectfont}}, color={rgb,1:red,0.0;green,0.0;blue,0.0}, draw opacity={1.0}, rotate={0.0}}, xmajorgrids={false}, xmin={-0.6000000000000014}, xmax={20.6}, xticklabels={{}}, xtick={{}}, xtick style={draw=none}, xticklabel style={font={{\fontsize{8 pt}{10.4 pt}\selectfont}}, color={rgb,1:red,0.0;green,0.0;blue,0.0}, draw opacity={1.0}, rotate={0.0}}, x grid style={color={rgb,1:red,0.0;green,0.0;blue,0.0}, draw opacity={0.1}, line width={0.5}, solid}, axis x line*={left}, separate axis lines, x axis line style={{draw opacity = 0}}, scaled y ticks={false}, ylabel={}, y tick style={color={rgb,1:red,0.0;green,0.0;blue,0.0}, opacity={1.0}}, y tick label style={color={rgb,1:red,0.0;green,0.0;blue,0.0}, opacity={1.0}, rotate={0}}, ylabel style={at={(ticklabel cs:0.5)}, anchor=near ticklabel, at={{(ticklabel cs:0.5)}}, anchor={near ticklabel}, font={{\fontsize{11 pt}{14.3 pt}\selectfont}}, color={rgb,1:red,0.0;green,0.0;blue,0.0}, draw opacity={1.0}, rotate={0.0}}, ymajorgrids={false}, ymin={0.35618046155633776}, ymax={0.7659114545145511}, yticklabels={{}}, ytick={{}}, ytick style={draw=none}, yticklabel style={font={{\fontsize{8 pt}{10.4 pt}\selectfont}}, color={rgb,1:red,0.0;green,0.0;blue,0.0}, draw opacity={1.0}, rotate={0.0}}, y grid style={color={rgb,1:red,0.0;green,0.0;blue,0.0}, draw opacity={0.1}, line width={0.5}, solid}, axis y line*={left}, y axis line style={{draw opacity = 0}}, colorbar={false}]
    \addplot[color={rgb,1:red,0.0;green,0.0;blue,0.0}, name path={832575b1-b2b6-446e-aa47-8001a8523ca9}, draw opacity={1.0}, line width={1}, solid]
        table[row sep={\\}]
        {
            \\
            0.0  0.4823432601040162  \\
            0.8333333333333334  0.4616003052225949  \\
            1.6666666666666667  0.45385454544295467  \\
            2.5  0.44541283961663963  \\
            3.3333333333333335  0.44081169479489796  \\
            4.166666666666667  0.4121605009713489  \\
            5.0  0.40304304009322667  \\
            5.833333333333333  0.3959550425713564  \\
            6.666666666666667  0.3952495278072296  \\
            7.5  0.36777662173440046  \\
            8.333333333333334  0.40802452589304056  \\
            9.166666666666666  0.4461731626617858  \\
            10.0  0.5224955721757875  \\
            10.833333333333334  0.4677833720651737  \\
            11.666666666666666  0.4682796706089865  \\
            12.5  0.41228193079681785  \\
            13.333333333333334  0.4209019560732842  \\
            14.166666666666666  0.4840165941272772  \\
            15.0  0.48621059614021617  \\
            15.833333333333334  0.5534786355466049  \\
            16.666666666666668  0.6513833625898784  \\
            17.5  0.6832508192526212  \\
            18.333333333333332  0.7104365711818745  \\
            19.166666666666668  0.7358904662784639  \\
            20.0  0.7543152943364885  \\
        }
        ;
\end{axis}
\end{tikzpicture}

}

\newsavebox{\posteriorplot}
\sbox{\posteriorplot}{%
\input{paperplots/flowchart_posterior.tikz}
}

\newsavebox{\priorplot}
\sbox{\priorplot}{%
\input{paperplots/flowchart_prior.tikz}
}

\newsavebox{\firstprior}
\sbox{\firstprior}{%
\input{paperplots/flowchart_prior_1.tikz}
}
\newsavebox{\secondprior}
\sbox{\secondprior}{%
\input{paperplots/flowchart_prior_2.tikz}
}

\begin{figure}
\begin{center}
\begin{adjustbox}{max width=\linewidth}
\begin{tikzpicture}[
    node distance=1cm and 1.2cm,
    nodes={align=center},
    ortho/install shortcuts]
  \node[basic box=redviolet, header=Data Sample \(x_{t_i} \sim p(\cdot)\), anchor=north]
    (data) {
        \scalebox{0.5}{
            \usebox{\dataplot}
        }
    };
  \node[basic box=munsellblue, header=Posterior Samples \(z_{t_i} \sim q(\cdot | x_{t_i})\), anchor=north, below of=data, shift=(down:y_node_dist*3.5)]
    (posterior) {
        \scalebox{0.5}{
            \usebox{\posteriorplot}
        }
    };
    
  \node[basic box=onyx, header=Prior Samples \(y_{t_i} \sim q(\cdot)\), anchor=north, below of=posterior, shift=(down:y_node_dist*3.5)]
    (prior) {
        \scalebox{0.5}{
            \usebox{\secondprior}
        }
    };

  \coordinate (midpoint1) at ($(data.south east)!0.5!(posterior.north east)$);
  \coordinate (midpoint2) at ($(posterior.south east)!0.5!(prior.north east)$);

  \node [basic box=onyx, left=20pt of (data.south west)(posterior.north west)]
    (encode) {Encoder\\Input: \(x_{t_i}\)\\Output: \({\color{munsellblue}\phi}\)};
  \node [basic box=maxpurple]
    (sample) at ($(midpoint1) + (80pt, -10pt)$) {Likelihood \(p(x_{t_i} | z_{t_i})\)};
  \node [basic box=maxpurple]
    (kldiv) at ($(midpoint2) + (80pt, 9pt)$) {KL divergence\\ \(D_{KL}(z_{t_i} || y_{t_i})\)};
    
  \node [basic box=onyx, right=10pt of (data.east), shift=(down:0.5)]
    (projector0) {Projector};
  \node [basic box=onyx, right=10pt of (posterior.east)]
    (projector1) {Projector};
  \node [basic box=onyx, right=10pt of (prior.east), shift=(up:0.5)]
    (projector3) {Projector};
    
  \path[->]
    (data.west) edge[->, ultra thick, out=180, in=90] (encode.north)
    (encode.south) edge[->, ultra thick, out=270, in=180] (posterior.west)

    (data.east) + (0,-0.5)edge[->, ultra thick] (projector0.west)
    (posterior.east) + (0,0) edge[->, ultra thick] (projector1.west)
    (prior.east) + (0,0.5) edge[->, ultra thick] (projector3.west)

    (projector0.east) edge[->, ultra thick, in=90, out=0] ($(sample.north) + (0.0,0)$)
    (projector1.east) edge[->, ultra thick, in=270, out=0] ($(sample.south) + (0.0,0)$)
    (projector1.east) edge[->, ultra thick, in=90, out=0] ($(kldiv.north) + (0.0,0)$)
    (projector3.east) edge[->, ultra thick, in=270, out=0] ($(kldiv.south) + (0.0,0)$)

    ;
  \end{tikzpicture}
\end{adjustbox}
\end{center}
\caption{Sketch of the latent neural SDE approach. The data sample is encoded and given as the context \(\color{munsellblue}\phi\) to the posterior SDE, which yields a probability distribution. The likelihood of the data given realizations of the distribution is maximized during training. The distance between the posterior SDE and the prior SDE is quantified by the KL divergence, which is minimized during training.} 
\label{fig:setup}
\end{figure}

In this section, we review latent neural SDEs.
Latent neural SDEs, as introduced in \citealt{li2020scalable}, are generative models that aim to reproduce the probability distribution of a data set consisting of time series. 
A latent neural SDE may be seen as consisting of two neural SDEs: a so-called posterior and a prior SDE.
The names \enquote{prior} and \enquote{posterior} stem from the variational inference framework. In this context, the posterior is approximate.
Each of these SDEs consists of a drift and a diffusion term as in Equation \eqref{eq:SDE}.
We first provide an intuitive overview and then explain the components that form the latent neural SDE.

\subsection{Intuitive Overview}
Latent neural SDEs aim to match the distribution of the \emph{prior SDE} to the data distribution. Because the distribution of the prior SDE and the data set are not easily comparable, this comparison is mediated via a second SDE, which is the \emph{posterior SDE} (cf.~\Cref{fig:setup}).

Latent neural SDEs aim to align the prior SDE's distribution to the posterior SDE's distribution and, at the same time, the posterior SDE's distribution to the data distribution. 
The prior SDE's distribution is aligned with the posterior SDE's distribution by minimizing the \emph{Kullback-Leibler (KL) divergence} between the two SDEs. 
The posterior SDE's distribution is aligned with the data distribution by feeding a \emph{context}, which is derived from samples of the data, into the posterior SDE.
The posterior SDE is then trained to reconstruct these samples. 

\subsection{Prior SDE and Posterior SDE}
The posterior SDE is a neural SDE of the form
\begin{align*}
  d u(t) &= h_{{\color{redviolet}\theta}}(u(t), t, {\color{munsellblue}\phi}) dt + g_{\color{redviolet}\theta}(u(t), t) d B_t \tag{posterior}
\end{align*}
where \({\color{redviolet}\theta}\) are the neural SDE's parameters, and \({\color{munsellblue}\phi}\) is the context, i.e., time-dependent information that is computed by another neural network, using an input trajectory from the data set. 
Given \({\color{munsellblue}\phi}\), the posterior SDE induces a probability distribution that aims to reconstruct the data distribution.

This reconstruction of the data by the posterior SDE is scored by an \emph{observation model}. The observation  model computes the likelihood that the data trajectories occur around the posterior SDE's trajectories, which is maximized during training. In latent neural SDEs, we manually define a Gaussian prior with a fixed variance around the posterior SDE's points and quantify the likelihood over this prior. Following \citealt{li2020scalable}, we write in integral form:
\[
  \mathcal{L}_E = \int_{0}^{t_{\text{train}}} \log p(x_{t} | z_{t}) dt\,,
\]
where $x_t$ denote data samples and $z_t$ sampled from integrating the posterior SDE. We illustrate the reconstruction that the posterior performs in \cref{fig:setup}. In training, we draw samples from the posterior SDE's distribution and minimize their distance to the input data using the likelihood.

The context \({\color{munsellblue}\phi}\) can be computed by a recurrent neural network (RNN), which we call the \textit{encoder}. The RNN operates on the reversed input trajectory, so its output at time \(t\) contains information from the input trajectory \(t\) until \(t_{\text{train}}\). This output at time \(t\) is an input to the posterior SDE's drift at time \(t\). This way, the posterior SDE has access to relevant information about the input trajectory's future.

The posterior alone cannot model the data's probability distribution: it always needs data as input (through the context) and thus does not provide a generative model on its own. 
Additionally, if we were to train the posterior alone, its diffusion would eventually be zero: The observation model would attempt to find a drift term that is close to the data by repeating the next value at each time step, observed through the context.

The prior SDE aims to model the data distribution without further input. Its general form is
\begin{align*}
  d \tilde{u}(t) &= f_{\color{redviolet}\theta}(\tilde{u}(t), t) dt + g_{\color{redviolet}\theta}(\tilde{u}(t), t) d B_t \tag{prior}
\end{align*}
which lacks the context \({\color{munsellblue}\phi}\). Note that while its drift $f$ is different from the posterior drift $h$, the prior and posterior share the same diffusion $g$. Because of this, we can quantify the KL divergence between both SDEs \citep{li2020scalable, tzen2019neural}. Importantly, this is the KL divergence of the distributions over the path space, not the KL divergence over the marginals.

\subsection{Encoder and Projector}
The encoder transforms a given path from data into the context \({\color{munsellblue}\phi}\). Theoretically, the encoder is not necessary, as we could directly set \({\color{munsellblue}\phi} = x_{t_i}\). However, choosing the encoder as a recurrent neural network that has seen future observations in the data has proven helpful \cite{li2020scalable,kidgerthesis}. The intuitive advantage of an encoder is that the posterior SDE can look as far into the future as needed.

The latent space (containing \(y_{t_i}\) and \(z_{t_i}\)) and data space (containing \(x_{t_i}\)) might differ. For instance, the latent space could have more dimensions than the data space. In this case, we need a \emph{projector} in the architecture, which would usually be a simple feed-forward neural network transforming a point in the latent space to a point in the data space.

\subsection{Training Objective}
With prior and posterior as above, the KL divergence between prior and posterior is
\begin{align*}
  \mathcal{L}_{KL} &= D_{KL}(\mu_q || \mu_p) \\
  &= \mathbb{E}_{B_t} \left[\int_0^{t_{\text{train}}} \frac{1}{2} \left\lVert \frac{ h_{
    {\color{redviolet}\theta}}(u(t), t, {\color{munsellblue}\phi}) -
  f_{\color{redviolet}\theta}(u(t), t) }{ g_{\color{redviolet}\theta}(u(t), t) } \right\rVert^2_2 dt\right].
\end{align*}
Here, \(\mu_q\) is the probability distribution of the posterior SDE, and \(\mu_p\) of the prior SDE, and we integrate along a trajectory \(u(t)\) of the posterior SDE along Brownian motion \(B_t\) \citep{li2020scalable}.

We jointly minimize the distance between the data distribution and the posterior SDE's distribution, as well as the distance between the prior SDE's and the posterior SDE's distribution, yielding the objective
\[
  \max_{\theta, \phi} \mathbb{E}_{x_{\text{data}}} (\mathcal{L}_E - \mathcal{L}_{KL}).
\]
This objective can be interpreted as a so-called evidence lower bound (ELBO) from variational inference, making a latent neural SDE a Bayesian variational autoencoder, as first proposed by \citealt{autoencoding}.

To have more control over the model, it is desirable to weigh the KL-divergence using a tunable factor \(\beta\) \citep{higgins2016beta, li2020scalable}: \[ \max_{\theta, \phi} \mathbb{E}_{x_{\text{data}}} (\mathcal{L}_E - \beta \mathcal{L}_{KL}). \] We explore the selection of this hyperparameter in the section below.

\section{Experimental Setup}
\label{sec:experiments}

\begin{table}
\centering
\begin{tabular}{lll}\toprule
\textbf{Aspect} & \textbf{Li et al.} & \textbf{Our setup} \\ \colrule
Diffusion Activation & Sigmoid & Sigmoid and softplus \\ 
Latent Space Dims.\ & More than data & Matches data \\ 
Projector & Trainable & None \\ 
Training & Adjoint SDE& Autodifferentiation \\ 
\botrule
\end{tabular}
\caption{Differences between Li et al.'s setup and ours}
\label{tab:comparison}
\end{table}

In the following we will consider two experimental systems.  
First, we set up a simple bistable stochastic system inspired by energy balance models (EBMs) in climate dynamics. The drift of this model exhibits two stable fixed points and the diffusion enables transitions between these two fixed points and their basins of attraction. Its governing equation is given by  
\[
    dT(t) = (a_1 + a_2 \tanh (T(t) - T_0) - a_3 T(t)^4) dt + \sigma d B_t
\]
where \(a_1 = 235.175\), \(a_2 = 81.8\), \(a_3 = 3.402\), \(T_0 = 273\), \(\sigma=40\).
These constants are similar to those found in the literature \citep{fraedrich1979catastrophes, ghil1976climate, sutera}. The EBM enables us to analyse the noise estimation of latent neural SDEs with a simple intuitive observable: the transition rate, which we define as the frequency at which trajectories cross the unstable fixed point between the two stable fixed points of the drift of the EBM.
In a one-dimensional SDE, back-and-forth transitions over the unstable fixed point can only be caused by diffusion. Thus, assuming the drift is estimated correctly, the transition rate is a measure of whether the diffusion size is estimated correctly.

In the EBM, we have slightly increased values \(\alpha_1\) and \(\alpha_2\) from a realistic model to yield closer fixed points for more frequent transition behavior. 
Using the EBM, we intend to test whether latent neural SDEs can a) model multistable systems and b) accurately fit the diffusion size and, therefore, the rate of transition of such systems.
We generally follow \citealt{li2020scalable} for our explicit latent neural SDE setup, except for the choices below. The differences are summarized in \cref{tab:comparison}.

\subsection{Latent Neural SDE Architecture}
We use gradients computed by directly autodifferentiating through the SDE solver, as opposed to solving an adjoint SDE, and choose an Euler-Maruyama SDE solver.

\citealt{li2020scalable} add a sigmoid activation after the final layer of the neural network of the diffusion term, and thus the value of the diffusion term is limited to the range \([0, 1]\).
In our experiments, the diffusion did often train to the value 1 in this setup. 
To remove the limit on the diffusion value while keeping the sigmoid's limiting effect, we add another layer with a final softplus activation to keep the output positive. 
We also attempted to leave out the sigmoid activation altogether, but this led to instabilities while training.

As in \citealt{li2020scalable}, the prior and posterior SDEs are time-independent, so they do not receive the current time \(t\) as input, only the current position in state space \(u(t)\). The encoder is also time-independent, so it does not receive time information alongside the data.

We use a one-dimensional latent space for data generated from a model with a one-dimensional state space. 
In contrast, \citealt{li2020scalable} use a four-dimensional latent space for data generated from a model with a two-dimensional state space. They use a trainable projector, whereas we use no projector to make the model more transparent, which in turn enables our analysis of noise underestimation of the unmodified latent neural SDE. 

\subsection{Training and Test Set}
The model trains on the timespan \([0, t_{\text{train}})\), which can be different for each model. We set \(t_{\text{train}} = 4\) time units for the EBM. 
The integration step size for the EBM is \(\Delta t = 0.01\).
To evaluate the statistics of the trained model, we use \([0.5 t_{\text{train}}, t_{\text{train}})\) as the training set and \([t_{\text{train}}, 5 t_{\text{train}})\) as the test set because we are less interested in the evolution from the Gaussian initial state and more interested in the long-run behavior of the system.

\subsection{Data Preprocessing and Fixed Hyperparameters}
We normalize the data by subtracting the mean and dividing by the standard deviation. We do not corrupt or perturb the data after generation.
We use an observation model based on a log-likelihood and we choose a variance of \(0.01\) following \citealt{li2020scalable}. Changing the variance will impact the optimal values for \(\beta\).
We always use the following hyperparameters: we train for 10000 epochs and use a linear annealing schedule for the KL divergence \citep{fu2019cyclical} that ramps up beta from zero to its final value in 1000 epochs; training without such a schedule is less stable in our experience. We use ADAM \cite{kingma2014adam} as the optimizer with a learning rate that starts at \(0.01\) and decays by a factor of \(0.997\) each iteration. Our batch size is 1024, and we use two hidden layers with 100 neurons in all networks.

\subsection{Analysis of Results}
We compute the marginals as histograms for the prior SDEs and the data. To quantify how close these are numerically, we use the \emph{Wasserstein distance}, which intuitively represents the cost of transforming one distribution into another \citep{panaretos2019statistical} and the \emph{transition rate}, which is the rate at which trajectories leave the area of influence of one attractor and tip into the other. 
Additionally, we directly compare the drift and diffusion of some trained latent neural SDEs with the EBM. 
Not only do we directly compare the drift and diffusion terms of both the ground truth system and the trained prior SDE, but we also compute and compare the first two Kramers-Moyal coefficients using the implementation of \citealt{gorjao2019kramersmoyal}.
The first two Kramers-Moyal coefficients are quantification of an SDE's drift and diffusion terms given data output by the system. 
We compute these factors even though we know the real drift and diffusion; our purpose is that different SDEs (i.e., the original EBM and the prior SDE) may produce similar behavior represented in different ways, resulting in more similar KM coefficients than their actual drift and diffusion terms.

\begin{figure}
    \centering
    \begin{subfigure}{.49\linewidth}
        \centering
        \begin{adjustbox}{width=\linewidth}
        \input{paperplots/betas/marginals_5_extrapolation_001.pgf}
        \end{adjustbox}
        \caption{Marginals, \(\beta=0.01\)}
    \end{subfigure}
    \begin{subfigure}{.49\linewidth}
        \centering
        \begin{adjustbox}{width=\linewidth}
        \input{paperplots/betas/marginals_5_extrapolation_1.pgf}
        \end{adjustbox}
        \caption{Marginals, \(\beta=1\)}
    \end{subfigure}
    \begin{subfigure}{.49\linewidth}
        \centering
        \begin{adjustbox}{width=\linewidth}
        \input{paperplots/betas/marginals_5_extrapolation_100.pgf}
        \end{adjustbox}
        \caption{Marginals, \(\beta=100\)}
    \end{subfigure}
    \begin{subfigure}{.49\linewidth}
        \centering
        \begin{adjustbox}{width=\linewidth}
        \input{paperplots/betas/marginals_5_extrapolation_10000.pgf}
        \end{adjustbox}
        \caption{Marginals, \(\beta=10000\)}
    \end{subfigure}
    \caption{Marginals for different values of hyperparameter \(\beta\) on test set}
    \label{fig:first_marginals}
\end{figure}

\begin{figure}
    \begin{subfigure}{\linewidth}
        \begin{center}
            \begin{adjustbox}{width=\linewidth}
        \input{paperplots/betas/custom_wasserstein.pgf}
            \end{adjustbox}
        \end{center}
        \caption{Wasserstein distance of marginals}
        \label{fig:first_wasserstein}
    \end{subfigure}
    \begin{subfigure}{\linewidth}
        \begin{center}
            \begin{adjustbox}{width=\linewidth}
        \input{paperplots/betas/custom_tipping.pgf}
            \end{adjustbox}
        \end{center}
        \caption{Transition rate}
        \label{fig:first_tipping_rate}
    \end{subfigure}
    \caption{Comparison of Wasserstein distance and transition rate between trained prior SDE and EBM for different values of hyperparameter \(\beta\) on training set and test set}
\end{figure}

\subsection{Selection of KL Weighing Factor}

The main hyperparameter to vary in a latent neural SDE is the KL weighting factor \(\beta\). We sweep \(\beta\) from \(0.01\) to \(10000\) in increments of a factor \(10\). 

We train on the EBM data and plot the marginals in \cref{fig:first_marginals}. 
There seems to be a sweet spot for \(\beta\) between \(10^1\) and \(10^3\), where the Wasserstein distances of the marginals between data and prior SDE are much lower. For \(\beta\) smaller or larger, the histograms of the latent neural SDEs's prior become more concentrated on the attractors and less broadly around them.

The Wasserstein distances of the train and test sets on the EBM data set, depending on the hyperparameter \(\beta\), are shown in \cref{fig:first_wasserstein}. The Wasserstein distances of the test set are close to those of the training set because we chose a one-dimensional latent neural SDE with the knowledge that the underlying model is one-dimensional. 

Next, we investigate the transition rates in \cref{fig:first_tipping_rate}. We can see that the transition rates of all latent neural SDEs are too low, less than half of the EBM's transition rates. 
With values of \(\beta\) that are too high or too low, the transition rates decrease alongside the size of the diffusion. Underestimation of the diffusion size is a known problem of latent neural SDEs, mentioned by \citealt{li2020scalable} and subsequently discussed in GitHub issues of packages that implement them. However, there is, as of now, no answer to why the model behaves this way. In the next section, we investigate this effect.

\section{Investigation of Noise Underestimation}\label{sec:investigation}

We have observed that latent neural SDEs underestimate the size of the diffusion. To understand the role of diffusion in the training procedure of latent neural SDEs, recall the loss function of a latent neural SDE
\begin{align*}
  &\max_{\theta, \phi} \; \mathbb{E}_{x_{\text{data}}} (\mathcal{L}_E - \beta \mathcal{L}_{KL}),
  \text{ where} \\ &\mathcal{L}_{KL} = \mathbb{E}_{B_t} \left[\int_0^{t_{\text{train}}} \frac{1}{2} \left\lVert \frac{ h_{
    {\color{redviolet}\theta} }(u(t), t, {\color{munsellblue}\phi}) -
  f_{\color{redviolet}\theta}(u(t), t) }{ g_{\color{redviolet}\theta}(u(t), t) } \right\rVert^2_2 dt\right].
\end{align*}
The size of the diffusion results from a balance of the two scores: the negative KL divergence \(-\mathcal{L}_{KL}\) correlates to the size of the diffusion, as long as the distance between the prior and posterior SDE's distributions is not zero, and the likelihoods \(\mathcal{L}_E\) correlate inversely with the size of the diffusion because the highest likelihoods can be achieved with zero diffusion.

The weighing factor \(\beta\) is the key to tuning the balance between the two scores, and thus, we reconsider the transition rate of the trained prior SDE for different values of \(\beta\) in \cref{fig:first_tipping_rate} in light of this balance.

For very small values of \(\beta\), the magnitude of the trained diffusion is close to zero, as only the posterior loss \(\mathcal{L}_E\) is minimized. This is reflected in the transition rate. With zero diffusion, the posterior SDE can trace the data sample with the highest accuracy. In this extreme case, the diffusion size of the latent neural SDE is independent of the amount of noise in the data.

The transition rate significantly decreases with high values of \(\beta\). This is surprising when considering the formulation of the training loss: as the KL divergence is proportional to the inverse of the diffusion and \(\beta\) increases its importance, we expected the size of the diffusion term to continue to increase with increasing \(\beta\). A possible explanation is that as \(\beta\) increases so much that the log-likelihoods become unimportant, there is no need to increase the diffusion if the prior and posterior SDEs' distributions are so close that their difference is almost zero. An extreme case of this is that the prior and posterior are equal, making the size of the diffusion arbitrary. We experimentally test this argument in \cref{appendix:argument}.

\begin{figure}
    \begin{subfigure}{\linewidth}
    \begin{center}
        \begin{adjustbox}{width=\linewidth}
    \input{paperplots/investigation/custom_wasserstein.pgf}
        \end{adjustbox}
    \end{center}
    \caption{Wasserstein distance (NaN in last experiment)}
    \label{fig:noise_wasserstein}
    \end{subfigure}
    \begin{subfigure}{\linewidth}
    \begin{center}
        \begin{adjustbox}{width=\linewidth}
    \input{paperplots/investigation/custom_tipping.pgf}
        \end{adjustbox}
    \end{center}
    \caption{Transition rate}
    \label{fig:noise_tipping_rate}
    \end{subfigure}
    \caption{Comparison of Wasserstein distane and transition rate between trained prior SDE and EBM for different EBM diffusion factors \(\sigma\) on training set and test set}
\end{figure}

\begin{figure}
    \centering
    \begin{subfigure}{.49\linewidth}
        \centering
        \begin{adjustbox}{width=\linewidth}
        \input{paperplots/investigation/training_info_data_noise_level_kl.pgf}
        \end{adjustbox}
        \caption{KL divergence for EBM \(\sigma\)}
    \end{subfigure}
    \begin{subfigure}{.49\linewidth}
        \centering
        \begin{adjustbox}{width=\linewidth}

        \input{paperplots/investigation/training_info_data_noise_level_logpxs.pgf}
        \end{adjustbox}
        \caption{Log-likelihood for EBM \(\sigma\)}
    \end{subfigure}
    \begin{subfigure}{\linewidth}
        \centering
        \begin{adjustbox}{width=0.5\linewidth}
        \input{paperplots/investigation/training_info_data_noise_level_noise.pgf}
        \end{adjustbox}
        \caption{Diffusion size of latent neural SDE for EBM diffusion factors \(\sigma\), integrated over training timespan}
    \end{subfigure}
    \caption{Training information for latent neural SDE for different EBM diffusion factors \(\sigma\): KL divergence, log-likelihood, and diffusion size (the diffusion size of the latent neural SDE is in the normalized space, while the diffusion size of EBM is in non-normalized space)}
    \label{fig:noise_kl_logpxs}
\end{figure}

\begin{figure}
    \centering
    \begin{subfigure}{.49\linewidth}
        \centering
        \begin{adjustbox}{width=\linewidth}
        \input{paperplots/investigation/diffusion_balance_2_train_0.pgf}
        \end{adjustbox}
        \caption{Diffusion balance, EBM diffusion \(\sigma=0\)}
    \end{subfigure}
    \begin{subfigure}{.49\linewidth}
        \centering
        \begin{adjustbox}{width=\linewidth}
        \input{paperplots/investigation/diffusion_balance_2_train_100.pgf}
        \end{adjustbox}
        \caption{Diffusion balance, EBM diffusion \(\sigma=100\)}
    \end{subfigure}
    \caption{Replaced actual trained latent neural SDE's diffusion with constant value \(\sigma_{\text{latent}}\) and computed both components of loss function. The latent neural SDE trains to align to the values of \(\sigma_{\text{latent}}\) where the scores, as weighted in this figure, have the smallest sum}
    \label{fig:balance}
\end{figure}

For other values of \(\beta\), it is unclear how the data noise and the latent neural SDE's diffusion size relate at first glance. We check if they have any relationship by varying the diffusion size \(\sigma\) of the EBM (c.f. \cref{sec:experiments}) and training a latent neural SDE with \(\beta=10\) on the resulting data. The results are shown in \cref{fig:noise_wasserstein} and \cref{fig:noise_tipping_rate}, where \enquote{EBM diffusion factor \(\sigma\)} is the size of the constant in the linear diffusion term of the EBM that is generating the data. We observe a correspondence between \(\sigma\) and the size of the latent neural SDE's diffusion, although the latent neural SDE's diffusion is always smaller.

It is more enlightening to plot the size of the KL divergence, log-likelihood, and diffusion of the latent neural SDE, trained on EBM data with different values of the diffusion factor \(\sigma\), as in \cref{fig:noise_kl_logpxs}. First, we observe that the data gets harder to approximate with increasing EBM diffusion factors \(\sigma\), as both the KL divergence and log-likelihood scores worsen (i.e., the KL divergence increases and the log-likelihoods decrease). The latent neural SDE's diffusion size grows almost linearly with \(\sigma\), similar to the KL divergence and log-likelihood.

This can be explained as follows: For latent neural SDEs, there is a \enquote{balance point} between the KL divergence and the log-likelihood. The latent neural SDE can freely choose the diffusion size according to these scores while training, and this can be seen in \cref{fig:noise_kl_logpxs}. As such, the diffusion size is a quantification of the \enquote{uncertainty} of the latent neural SDE's drift about the data drift, which is not necessarily the data's diffusion size.
We add that this behaviour is perfectly acceptable in uncertainty quantification, where the diffusion size is a measure of the uncertainty of the model. However, in the context of the EBM, the diffusion size is a measure of the noise in the data, and the latent neural SDE should be able to model this noise.

How is the diffusion size chosen in training? A larger diffusion size makes for a better KL divergence and a worse log-likelihood, and a smaller diffusion size makes for a better log-likelihood and worse KL divergence. To demonstrate this, we now use the latent neural SDE for \(\beta=10\) and subsequently replace the trained diffusion of the latent neural SDE with constant values from zero to five in \cref{fig:balance}. We plot both terms of the loss function for two data diffusion factors \(\sigma\), zero and 100.

Here, the actual size of the final model's diffusion turns out to be about the value of \(\sigma\) where these two scores, as weighted in the figure, have the smallest sum (i.e., the latent neural SDE's objective is maximized). The harder the problem becomes to predict, the further this point moves to the right, leading to the relationship between the EBM's diffusion factors \(\sigma\) and the latent neural SDE's diffusion size we observed.

\section{Injecting Noise Into Latent neural SDEs}
\label{sec:noise}

In order to match the transition rate as well as the marginals, we amend the loss function to include a term for the size of the diffusion:
\begin{align*}
    \max_{\theta, \phi} \mathbb{E}_{x_{\text{data}}} (\mathcal{L}_E - \beta \mathcal{L}_{KL} + \gamma \mathcal{L}_G) \\
    \text{where } \mathcal{L}_G = \int_0^T \left\lVert g_{\theta}(u(t), t) \right\rVert dt.
\end{align*}
Alternatively, assuming we would know the desired value of \(\mathcal{L}_N\), we could use a regula\-ri\-za\-tion-like penalty:
\begin{align*}
    \max_{\theta, \phi} \mathbb{E}_{x_{\text{data}}} (\mathcal{L}_E - \beta \mathcal{L}_{KL} + \gamma (\mathcal{L}_G - g_{\text{target}})^2).
\end{align*}
However, we do not know \(g_{\text{target}}\) in general, so this adds two hyperparameters instead of one. Thus, we choose the first loss function. We compute \(\mathcal{L}_G\) just like \(\mathcal{L}_{KL}\), adding another dimension in the integration of the posterior \cite[cf.][]{li2020scalable}.

\begin{figure}
    \begin{subfigure}{\linewidth}
    \begin{center}
        \begin{adjustbox}{width=\linewidth}
    \input{paperplots/penalty/custom_wasserstein.pgf}
        \end{adjustbox}
    \end{center}
    \caption{Wasserstein distance}
    \label{fig:gamma_wasserstein}
    \end{subfigure}
    \begin{subfigure}{\linewidth}
    \begin{center}
        \begin{adjustbox}{width=\linewidth}
    \input{paperplots/penalty/custom_tipping.pgf}
        \end{adjustbox}
    \end{center}
    \caption{Transition rate}
    \label{fig:gamma_tipping_rate}
    \end{subfigure}
    \caption{Comparison of Wasserstein distance and transition rate between EBM and trained prior SDE for different values of noise penalty \(\gamma\) on training set and test set}
\end{figure}

\begin{figure}
    \begin{subfigure}{\linewidth}
        \begin{center}
        \begin{adjustbox}{width=\linewidth}
            \input{paperplots/penalty/prior_0_firsthalftrain_big.pgf}
        \end{adjustbox}
        \end{center}
        \caption{Noisy EBM}
    \end{subfigure}
    \begin{subfigure}{\linewidth}
        \begin{center}
        \begin{adjustbox}{width=\linewidth}
            \input{paperplots/penalty/data_0_firsthalftrain_big.pgf}
        \end{adjustbox}
        \end{center}
        \caption{Latent neural SDE Prior, \(\beta=10, \gamma=200\)}
    \end{subfigure}
    \caption{Comparison of best result with the latent neural SDE with noise penalty against noisy EBM}
    \label{fig:simulationsnoiseebm}
\end{figure}

\begin{figure}
    \begin{center}
    \begin{adjustbox}{width=\linewidth}
        \input{paperplots/penalty/marginals_5_extrapolation_big.pgf}
    \end{adjustbox}
    \end{center}
    \caption{Marginals on test set, \(\beta=10, \gamma=200\)}
    \label{fig:marginalsnoiseebm}
\end{figure}

\begin{figure}
    \begin{center}
        \begin{adjustbox}{width=\linewidth}
    \input{paperplots/penalty/kramersmoyal_noise_penalty_200_5_extrapolation.pgf}
        \end{adjustbox}
    \end{center}
    \caption{Comparison of the Kramers-Moyal factors (drift left, diffusion right) between EBM and trained prior SDE with noise penalty \(\gamma=0\) and \(\gamma=200\) on test set}
    \label{fig:gamma_km_factors}
\end{figure}

We start by fixing the balance factor \(\beta = 10\) and training a latent neural SDE on the EBM for different values of the new noise penalty \(\gamma\).
In \cref{fig:gamma_tipping_rate}, we observe that increasing \(\gamma\) will increase the transition rate. Relating the Wasserstein distance to \(\gamma\) in \cref{fig:gamma_wasserstein}, we cannot observe a clear pattern but see values similar to those that we would obtain by setting \(\gamma=0\) in the test set. The latent neural SDE trained with \(\gamma = 200\) matches the transition rate and Wasserstein distances quite well.

We emphasize that this process of finding the correct value of \(\gamma\) can be done without any knowledge of the underlying dynamics, just by analysing the data.
Furthermore, any metric can be chosen to optimize the noise penalty, and this metric will have to be fine-tuned for the use case, such as our choice of the transition rate for the EBM.

\begin{figure}
    \centering
    \begin{subfigure}{.49\linewidth}
        \centering
        \begin{adjustbox}{width=\linewidth}
        \input{paperplots/penalty/func_drift_0.pgf}
        \end{adjustbox}
        \caption{Drift, \(\gamma=0\)}
    \end{subfigure}
    \begin{subfigure}{.49\linewidth}
        \centering
        \begin{adjustbox}{width=\linewidth}
        \input{paperplots/penalty/func_drift_200.pgf}
        \end{adjustbox}
        \caption{Drift, \(\gamma=200\)}
    \end{subfigure}
    \centering
    \begin{subfigure}{.49\linewidth}
        \centering
        \begin{adjustbox}{width=\linewidth}
        \input{paperplots/penalty/func_diffusion_0.pgf}
        \end{adjustbox}
        \caption{Diffusion, \(\gamma=0\)}
    \end{subfigure}
    \begin{subfigure}{.49\linewidth}
        \centering
        \begin{adjustbox}{width=\linewidth}
        \input{paperplots/penalty/func_diffusion_200.pgf}
        \end{adjustbox}
        \caption{Diffusion, \(\gamma=200\)}
    \end{subfigure}
    \caption{Direct comparison of drift and diffusion functions of prior SDE and EBM for different values of the noise penalty \(\gamma\)}
    \label{fig:value_of_diffusion_gamma}
\end{figure}

Having found a seemingly reasonable value \(\gamma=200\) for the noise penalty by looking at the transition rates, we now investigate how well the trained latent neural SDE with \(\beta = 10\) and \(\gamma = 200\) fits the EBM. We plot simulations in \cref{fig:simulationsnoiseebm} and the marginals in \cref{fig:marginalsnoiseebm}. Next, we compare both the KM coefficients of drift and diffusion estimated from the data and the actual values of drift and diffusion between the latent neural SDE's prior SDE and the EBM.
While we do not perform a grid search to find the optimal combination of \(\gamma\) and \(\beta\), performing one is a good means to find the best possible parameterization.

In \cref{fig:gamma_km_factors}, we use the Epanechnikov kernel found in the \texttt{kramersmoyal} Python package \citep{gorjao2019kramersmoyal} to compare Kramers-Moyal coefficients (see also \cref{sec:appendix:km}).
We compute these from all data samples and latent neural SDEs with \(\gamma \in \{0, 200\}\) on the test set. The first and second coefficients represent the drift and diffusion, respectively. Here, the values from the latent neural SDE and the data look very similar.

Not only can we compute the KM coefficients from the data generated by the trained prior SDE and the EBM, but we can directly access their drift and diffusion functions and compare them. We do this for \(\gamma \in \{0, 200\}\) in \cref{fig:value_of_diffusion_gamma}. Here, we see quite a larger difference than with the KM coefficients. With noise penalty \(\gamma=200\), the latent neural SDE learns a drift and diffusion that is significantly different to the EBM's drift and diffusion, but these values of drift and diffusion yield trajectories that behave very similarly, as can be seen from the KM coefficients.

We conduct some further experiments to test the noise penalty in different scenarios:
In \cref{sec:appendix:fhn}, we demonstrate the effectiveness of our method on a more complex system, a two-dimensional FitzHugh-Nagumo (FHN) model that models the NGRIP ice core data \cite{lohmann2019consistent}.
In \cref{sec:appendix:ou}, we also train on a simple Ornstein-Uhlenbeck process, to show that the noise penalty also fixes the noise underestimation problem for a simple linear example.
In \cref{sec:appendix:triplewell}, we also train on a triple well model.
In \cref{sec:appendix:ebmrare}, we also train on an EBM with a lower level of noise that exhibits tipping more rarely.

We conclude that by setting the hyperparameter \(\gamma\) to the correct value w.r.t. the modelling goals, we can synthesize an SDE that matches Wasserstein distances and transition rates of the data relatively well in the considered examples with constant diffusion. In this setup, \(\gamma\) cannot be learned automatically, leaving us with two hyperparameters ($\beta$ and $\gamma$) that must be manually tuned. On the other hand, with enough data, correct values of these hyperparameters can be obtained just by analysing the data without any further knowledge of the underlying dynamics.

The hyperparameter \(\gamma\) only fits diffusion terms globally in its current formulation. The result is thus only accurate for data sets with constant diffusion. We show in \cref{sec:appendix:ebm} that our method fails if the noise of the data is linear. To fit more complicated diffusion functions, the loss term \(\mathcal{L}_G\) must be modified in some way to track the diffusion in more detail.

\section{Discussion}

We have shown that with the introduction of an additional noise penalty, latent neural SDEs can fit multistable dynamics, achieving high accuracy in KM coefficients, transition rate, and marginals. The noise penalty is essential to address diffusion underestimation in the stochastic models we considered, which include an energy balance model, FitzHugh-Nagumo model and an Ornstein-Uhlenbeck process model. We investigated the cause of the underlying noise underestimation effect, which lies in the formulation of the loss function for latent neural SDEs. The unmodified latent neural SDE has to achieve a balance between the KL divergence and the log-likelihood of the data, which in turn results in an underestimated diffusion size.

Our formulation of the noise penalty adds another tunable hyperparameter to the latent neural SDE. More sophisticated methods, such as automatic hyperparameter optimization, might be a successful approach for tuning the noise penalty automatically and thus automatically training a latent neural SDE to have the correct diffusion.

Using SDE-GANs to estimate these systems might have advantages, although the training procedure will be considerably more difficult. We expect that the noise underestimation problem would not arise for SDE-GANs in the same way, as the discriminator would ideally catch a trajectory with too little noise.
The latent neural SDE framework likely has modelling advantages over SDE-GANs, for instance, the possibility of prediction from data samples and out-of-distribution detection through the posterior SDE. One could give the posterior SDE an unseen data sample and make a forecast by switching to the prior after the last posterior point in the state space. Out-of-distribution detection might be possible by exploring the context returned by the encoder. To the best of our knowledge, this has not yet been attempted in the literature.

By modifying the latent neural SDE framework, we achieve a generative framework that is comparably easy to train and can fit multistable dynamics with an accurate estimation of their diffusion size. It thus significantly broadens the applications of the latent neural SDE method. 

\begin{acknowledgments}
N.B. acknowledges funding by the Volkswagen foundation. This is ClimTip contribution \#X; the ClimTip project has received funding from the European Union's Horizon Europe research and innovation programme under grant agreement No. 101137601.
M.T.S acknowledges funding by the Ministry of Culture and Science (MKW) of the German State of North Rhine-Westphalia (``NRW Rückkehrprogramm").
\end{acknowledgments}

\appendix

\begin{figure}
    \centering
    \begin{subfigure}{.49\linewidth}
        \centering
        \begin{adjustbox}{width=\linewidth}
        \input{paperplots/appendix/fhn_marginals_5_extrapolation.pgf}
        \end{adjustbox}
        \caption{Marginals, FHN, \(\beta=10, \gamma=500\)}
        \label{fig:marginalsfhn}
    \end{subfigure}
    \begin{subfigure}{.49\linewidth}
        \centering
        \begin{adjustbox}{width=\linewidth}
        \input{paperplots/ornstein/marginals_5_extrapolation.pgf}
        \end{adjustbox}
        \caption{Marginals, OU, \(\beta=10, \gamma=650\)}
        \label{fig:marginalsou}
    \end{subfigure}

    \begin{subfigure}{.49\linewidth}
        \centering
        \begin{adjustbox}{width=\linewidth}
        \input{paperplots/rare/marginals_5_extrapolation.pgf}
        \end{adjustbox}
        \caption{Marginals, EBM with rarer tipping, \(\beta=10, \gamma=150\)}
        \label{fig:marginalsrare}
    \end{subfigure}
    \begin{subfigure}{.49\linewidth}
        \centering
        \begin{adjustbox}{width=\linewidth}
        \input{paperplots/triplewell/marginals_5_extrapolation.pgf}
        \end{adjustbox}
        \caption{Marginals, Triple Well model, \(\beta=10, \gamma=205\)}
        \label{fig:triplewell}
    \end{subfigure}
    \caption{Marginals for experiments in appendix, all computed on test set}
    \label{fig:appendixmarginals}
\end{figure}

\section{Kramers-Moyal Coefficients}
\label{sec:appendix:km}

In this paper, we use the Kramers-Moyal (KM) coefficients to analyse the behaviour of our trained stochastic processes. The idea is that while we know the exact drift and diffusion terms of our ground truth and trained latent neural SDEs, they might behave more similarly than these suggest. Similar KM coefficients indicate this.

Given an \(N\)-dimensional Markovian process \(x(t) \in \mathbb{R}^n\), the Kramers-Moyal coefficients are given as
\begin{align*}
&\mathcal{M}^\sigma(x, t) = \\
&\lim_{\Delta t \rightarrow 0} \frac{1}{\Delta t}
\int
d x' [x'(t) - x(t)]^\sigma P(x', t + \Delta t \mid x, t),
\end{align*}
where \(P(x', t + \Delta t \mid x, t)\) is the transition probability density, square brackets denote dyadic multiplication and \(\sigma\) is a multi-index describing the orders \cite{gorjao2019kramersmoyal}.

\section{FitzHugh-Nagumo Model}
\label{sec:appendix:fhn}

\begin{figure}
    \begin{center}
        \begin{adjustbox}{width=\linewidth}
\input{paperplots/appendix/fhn.pgf}
        \end{adjustbox}
    \end{center}
    \caption{FitzHugh-Nagumo: One-dimensional KM coefficients (drift left, diffusion right) on test set}
    \label{fig:gamma_km_factors_fhn}
\end{figure}

To show the effectiveness of our method on a more complex system, we train a latent neural SDE on data generated by a two-dimensional monostable FitzHugh-Nagumo (FHN) model.
The FHN is a generalized Van der Pol oscillator designed initially as a simplified model for neuron spikes in nerves. Its state space has two dimensions. The original model has an attractor \(P\), which FitzHugh calls the resting point. When disturbed far enough from the \(P\), a trajectory will travel through the model's state space and arrive back at \(P\) at some point \cite{fitzhugh1961impulses}.

The governing equation of the FHN in its stochastic form is:
\begin{align*}
    d x(t) &= \frac{1}{\tau_x} (\alpha_1 x(t) - \alpha_3 x(t)^3 + y(t)) dt + \sigma_x dB_{x,t}, \\
    d y(t) &= \frac{1}{\tau_y} (\beta y(t) - x(t) + c) dt + \sigma_y dB_{y,t}.
\end{align*}
We choose an automatically synthesized parameterization from \citealt{lohmann2019consistent} that was selected to match certain behaviours of the NGRIP ice core record, which they call \(\text{FHN}_\gamma\),
where \(\tau_x = \tau_y = 1, b = 2.55, \alpha_1 = 0.63, \alpha_3 = 2.71, c = 0.22, \sigma_x = 4.80, \sigma_y = 11.08, \beta = \tan(-0.67)\).
Note that this system is monostable, and while it exhibits tipping-like dynamics, these are not due to the typical relaxation oscillation that the FHN is mostly known for.

We choose to observe just the first dimension of the FHN model in an analogy to the NGRIP ice core. Thus, we also demonstrate the capabilities of latent neural SDEs to model just partially observed nonlinear chaotic systems. 

We train a two-dimensional latent neural SDE on the data from this model, but only let it observe \(x(t)\), which is the dimension that models the NGRIP ice core record. We choose \(\beta=10\) and increase the learning rate decay to \(0.9997\).

By summary statistics, we select \(\gamma=500\) as the noise penalty.
We will restrict our comparative analysis of the fitted model to the first dimension \(x(t)\) of the FHN model, as this dimension models the NGRIP ice core record, and the latent neural SDE only observed this dimension.
We plot the KM coefficients for \(\gamma = 0\) and \(\gamma = 500\) in \cref{fig:gamma_km_factors_fhn}. The latent neural SDE's prior SDE is able to match the behaviour of the FHN model well with \(\gamma = 500\), but not as a standard latent neural SDE with \(\gamma = 0\).
The transition rate of the latent neural SDE with \(\gamma = 500\) is 513.29, which is close to the transition rate of the FHN model, 458.29, whereas for \(\gamma=0\) the transition rate is 232.4. The Wasserstein distances of both latent neural SDEs are similar (0.176 for \(\gamma=500\) and 0.206 for \(\gamma=0\)).
We show the marginals of the prior SDE with \(gamma=500\) in \cref{fig:marginalsfhn}.

\section{Minimal Ornstein-Uhlenbeck Process}
\label{sec:appendix:ou}

\begin{figure}
    \begin{center}
        \begin{adjustbox}{width=\linewidth}
\input{paperplots/ornstein/kramersmoyal_noise_penalty_650_5_extrapolation.pgf}
        \end{adjustbox}
    \end{center}
    \caption{Ornstein-Uhlenbeck process: KM coefficients (drift left, diffusion right) on test set}
    \label{fig:gamma_km_factors_ornstein}
\end{figure}

To show that the noise penalty also works with a minimal linear example, we train on an Ornstein-Uhlenbeck process of the form
\[
    du(t) = - u(t) dt + d B_t.
\]
The drift of this process is a simple attractor around zero, and the diffusion is one.
We chose \(t_\text{train} = 5\), \(\Delta t = 0.01\) and trained for 5000 epochs, keeping all other settings the same as for the EBM. We select \(\gamma=650\) by analyzing summary statistics and plot the KM coefficients for \(\gamma = 0\) and \(\gamma = 650\) in \cref{fig:gamma_km_factors_ornstein}. The latent neural SDE's prior SDE is able to match the behaviour of the Ornstein-Uhlenbeck process well with \(\gamma = 650\), but not as a standard latent neural SDE with \(\gamma = 0\).
We show the marginals of the prior SDE with \(\gamma=650\) in \cref{fig:marginalsou}.

\section{EBM With Rarer Tipping}
\label{sec:appendix:ebmrare}

\begin{figure}
    \centering
    \begin{subfigure}{.49\linewidth}
        \centering
        \begin{adjustbox}{width=\linewidth}
        \input{paperplots/rare/data_2_train.pgf}
        \end{adjustbox}
        \caption{Noisy EBM with rarer tipping}
    \end{subfigure}
    \begin{subfigure}{.49\linewidth}
        \centering
        \begin{adjustbox}{width=\linewidth}
        \input{paperplots/rare/prior_2_train.pgf}
        \end{adjustbox}
        \caption{Latent neural SDE Prior, \(\beta=10, \gamma=150\)}
    \end{subfigure}
    \caption{Comparison of best result with the latent neural SDE against noisy EBM with rarer tipping}
    \label{fig:simulationsrare}
\end{figure}

\begin{figure}
    \begin{center}
        \begin{adjustbox}{width=\linewidth}
\input{paperplots/rare/kramersmoyal_noise_penalty_150_5_extrapolation.pgf}
        \end{adjustbox}
    \end{center}
    \caption{EBM with rarer tipping: KM coefficients (drift left, diffusion right) on test set}
    \label{fig:kmfactorsextra}
\end{figure}

We repeat the same experiment as in \cref{sec:noise} with an EBM that exhibits tipping more rarely. For this, we keep all parameters the same, but we set \(\sigma = 25\) (in \cref{sec:noise}, we set it to \(40\)). We searched for \(\gamma\) as in \cref{sec:noise} and found that for \(\gamma = 150\), the transition rate is similar (47.8 for the trained prior SDE and 50.5 for the noisy EBM).
We plot the marginals in \cref{fig:marginalsrare} and the KM coefficients in \cref{fig:kmfactorsextra}.
We also show simulations in \cref{fig:simulationsrare}.

\section{Triple-Well Model}
\label{sec:appendix:triplewell}

\begin{figure}
    \begin{center}
        \begin{adjustbox}{width=\linewidth}
\input{paperplots/triplewell/kramersmoyal_noise_penalty_205_5_extrapolation.pgf}
        \end{adjustbox}
    \end{center}
    \caption{Triple-Well: One-dimensional KM coefficients (drift left, diffusion right) on test set}
    \label{fig:gamma_km_factors_triple_well}
\end{figure}

We have considered a double-well model so far. We now consider the following triple-well model:
\[
    dT(t) = -6 T(t)^5 + 20 T(t)^3 - 8T(t) dt + 2.5 d B_t.
\]
We train a one-dimensional latent neural SDE with \(\beta=10\). By looking at the KM coefficients, we select \(\gamma=205\). We plot the KM coefficients in \cref{fig:gamma_km_factors_triple_well} and the marginals in \cref{fig:triplewell}.

\section{EBM With Linear Diffusion}
\label{sec:appendix:ebm}

\begin{figure}
    \begin{center}
        \begin{adjustbox}{width=\linewidth}
    \input{paperplots/linear/kramersmoyal_noise_penalty_150_5_extrapolation.pgf}
        \end{adjustbox}
    \end{center}
    \caption{EBM with linear noise: KM coefficients (drift left, diffusion right) on test set}
    \label{fig:gamma_linear}
\end{figure}

The noise penalty \(\gamma\) globally increases the size of the latent neural SDE's diffusion function, so we expect our noise penalty method to fail to match diffusion functions that are not constant.
To test this hypothesis, we build an EBM with a linear diffusion term by slightly modifying the EBM equation in \cref{sec:experiments}:
\begin{align*}
    dT(t) =& (a_1 + a_2 \tanh (T(t) - T_0) - a_3 T(t)^4) dt + \\
    &0.135 T(t) d B_t.
\end{align*}
In this EBM, we have multiplicative noise, meaning that the diffusion depends on the value \(T(t)\). While our trained latent SDE could exhibit this behaviour, our formulation of the noise penalty does not take this into account.

We kept the training settings the same as for the original EBM. We select \(\gamma=150\) by analysing summary statistics and plot the KM coefficients for \(\gamma = 0\) and \(\gamma = 150\) in \cref{fig:gamma_linear}. While the drift's KM coefficients still match between the EBM and the latent neural SDE's prior with noise penalty, the diffusion is clearly linear for the EBM and close to constant for the latent neural SDE. Thus, one can indeed only use the noise penalty method to match constant diffusion functions. A noise penalty for multiplicative noise is an interesting avenue for future research.

\section{Latent Neural SDEs on Extreme Values of Beta}
\label{appendix:argument}

\begin{figure}
    \centering
    \begin{subfigure}{.49\linewidth}
        \centering
        \begin{adjustbox}{width=\linewidth}
            \input{paperplots/argument/training_info_Noise.pgf}
        \end{adjustbox}
        \caption{Noise level of latent neural SDE with \(\beta=0\) during training}
        \label{fig:noiselevel1}
    \end{subfigure}
    \begin{subfigure}{.49\linewidth}
        \centering
        \begin{adjustbox}{width=\linewidth}
            \input{paperplots/argument/training_info_Noise_2.pgf}
        \end{adjustbox}
        \caption{Noise level of latent neural SDE with \(\beta=10^{10}\) during training}
        \label{fig:noiselevel2}
    \end{subfigure}
    \caption{Noise level of latent neural SDEs}
\end{figure}
We aim to test our argument from \cref{sec:investigation} experimentally. For this, we train three latent neural SDEs on the EBM from \cref{sec:experiments}: all without noise penalty, one with \(\beta=0\), and two with \(\beta=10^{10}\). During the training of the latent neural SDE with \(\beta=0\), the noise level consistently moves toward zero, see \cref{fig:noiselevel1}. On the other hand, with \(\beta=10^{10}\), the noise level moves in the beginning but then stays almost constant, see \cref{fig:noiselevel2}. The two latent neural SDEs with \(\beta=10^{10}\) have different noise levels after training: one is around 3.2 while the other is around 3.8. Thus, the noise level indeed seems arbitrary here.

\bibliography{paper}

\end{document}